\documentclass{article}
\usepackage{spconf,amsmath,graphicx,hyperref}
\usepackage{cite}
\usepackage{url}
\usepackage{amsmath,amssymb,amsfonts}
\usepackage{algorithmic}
\usepackage{graphicx}
\usepackage{textcomp}
\usepackage{booktabs}
\usepackage{multirow}
\usepackage{xcolor}
\usepackage{makecell}
\usepackage[T2A,OT1]{fontenc}
\newcommand{\cyrtext}[1]{{\fontencoding{T2A}\selectfont #1}}
\usepackage{CJKutf8}
\usepackage[safe,tone]{tipa}

\makeatletter
\renewcommand\subsection{\@startsection{subsection}{2}{\z@}%
{-2ex \@plus -.5ex \@minus -.2ex}
{1.5ex \@plus .2ex}
{\normalfont\large\bfseries}}
\makeatother

\title{ETC-Bench: A Multilingual Benchmark for Evaluation-Side Transcript Canonicalization}
\name{Zhiheng Qian$^{1*}$, Junhao Du$^{1*}$, Jing Peng$^{1}$,
Hanqi Li$^{1}$, Kunyang Peng$^{2}$, Feng Xue$^{2}$, Shuai Wang$^{3}$,
Kai Yu$^{1\dagger}$\thanks{$^*$Equal contribution. $^\dagger$Corresponding author.\\
\hspace*{1.5em}\href{https://github.com/zh-qian2004/etc-bench}{https://github.com/zh-qian2004/etc-bench}}}
\address{$^1$X-LANCE Lab, School of Computer Science, Shanghai Jiao Tong University, Shanghai, China\\
$^1$MoE Key Lab of Artificial Intelligence, $^1$Jiangsu Key Lab of Language Computing\\
$^2$AISpeech Ltd, Suzhou, China \\
$^3$School of Intelligence Science and Technology, Nanjing University}
\begin{document}

\ninept
\maketitle
\begin{abstract}
Written transcripts of the same speech can differ in ways that are not uniquely determined by the speech signal. We formulate evaluation-side transcript canonicalization (ETC), which maps multiple surface forms to a single canonical form when their differences are not determined by speech. To our knowledge, ETC-Bench is the first and most comprehensive multilingual benchmark for evaluation-side transcript canonicalization, covering 181 phenomenon types across 12 languages. Each instance is a minimal pair that shares a sentence context and differs only in a target span exhibiting one phenomenon. To evaluate both canonicalization success and semantic preservation, we propose Semantic-preserving Exact-match Normalization Accuracy (SENA), which combines exact agreement between independently canonicalized pair members with a pair-relative BERTScore constraint against information loss. We further introduce the Taxonomy-guided Agent Canonicalization Skill (TACS), which provides LLM agents with phenomenon descriptions and explicit canonicalization contracts. Experiments with five baseline configurations and three TACS-based agents show that the agents outperform the baseline systems on shared languages. ETC-Bench provides a systematic testbed for evaluation-side transcript canonicalization across languages.
\end{abstract}

\begin{figure*}[t]
  \centering
  \includegraphics[width=\linewidth]{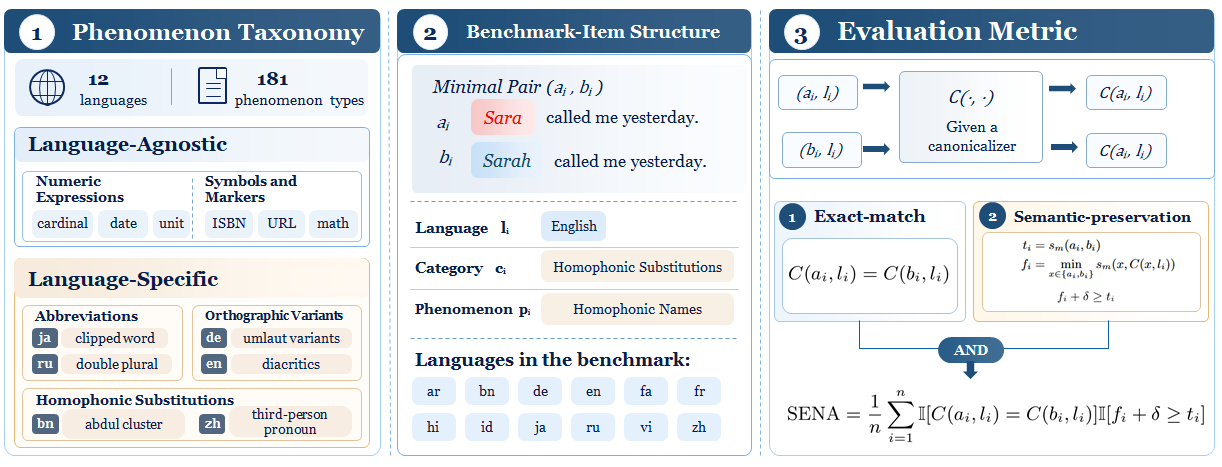}
  \caption{Overview of ETC-Bench: taxonomy, benchmark-item structure, and evaluation metric.}
  \label{fig:taxonomy}
\end{figure*}

\section{Introduction}
Speech-to-text (STT) systems \cite{peng2025survey,fang2025low} produce text from speech and are commonly evaluated by comparing hypotheses with references \cite{kim2021semdist}.However, speech does not always determine a unique written form.
Transcript ambiguity arises when a spoken utterance does not encode the graphemic or symbolic information needed to choose among alternative written forms with the same spoken realization.
For example, a number may be transcribed as ``2'' or ``two'', and a common
Chinese name pronounced ``/\textipa{sw@n\tone{55} li\tone{51}}/'' may have several
commonly used character forms, such as ``\begin{CJK*}{UTF8}{gbsn}孙丽\end{CJK*}'' and ``\begin{CJK*}{UTF8}{gbsn}孙立\end{CJK*}''.
Furthermore, in some languages the same word may have multiple spellings \cite{nigmatulina2020flexwer}.
Word error rate (WER) counts such differences as errors, conflating model performance with transcript conventions and potentially biasing cross-lingual comparisons \cite{srivastav2025openasr}.

Text normalization (TN) and inverse text normalization (ITN) are two major normalization paradigms in speech processing. They aim at the conversion between written and spoken forms \cite{sproat2001normalization,flint2017normalisation,deviyani2022normalization}.
TN maps written expressions to spoken forms, typically for text-to-speech (TTS) \cite{sproat2017rnn}, whereas ITN maps spoken-form text to conventional written text. 
NeMo provides weighted finite-state grammars \cite{gorman2016pynini} for TN and ITN classes such as numbers, dates, and measurements \cite{zhang2021nemo}. 
PolyNorm extends multilingual TN with an LLM-based pipeline and a written-to-spoken taxonomy \cite{wong2025polynorm}. 
Although TN and ITN were developed for different application needs, both can serve as canonicalization subroutines, depending on the chosen target representation.
In either case, applying the same mapping independently to each transcript can remove the surface distinction. However, existing TN and ITN systems cover only a limited subset of the ambiguities considered here. Many language-specific orthographic and homophonic variations fall outside the semiotic classes they normalize.

In practice, the evaluation of automatic speech recognition (ASR) already accommodates transcript variation in several ways to reduce the effect of this ambiguity to better measure model performance. 
Multi-reference evaluation accounts for acceptable spelling variation in Egyptian Arabic, but requires additional transcripts \cite{ali2015multireference}.
Sclite also accommodates transcript variation in its evaluation procedure \cite{fiscus2006multiple}. 
Whisper provides post-ASR text normalizers to reduce penalties for transcript variation \cite{radford2022whisper}, but its well-established normalization tool is limited to English. 
Furthermore, its inappropriate behavior can damage semantic information, as shown for Indic scripts \cite{manohar2024normalization}.
Despite these practices, multilingual evaluation-oriented normalization remains limited in scope. Recent speech-evaluation efforts, including SURE and Open-Bench, likewise emphasize reproducible pipelines and fair comparisons through standardized normalization and scoring \cite{peng2026unified,openbench}. Motivated by this principle, we define \emph{evaluation-side transcript canonicalization} (ETC) to map STT transcript variants to a consistent canonical form when their differences cannot be resolved from speech, supporting more trustworthy ASR evaluation and offering a framework relevant to speech-translation text outputs.
Given a transcript $x$ and its language $l$, a canonicalizer produces $C(x,l)$, without access to another transcript of the same utterance. 
Let $a\equiv_l b$ indicate that $a$ and $b$ are valid written representations of the same spoken utterance and differ only in distinctions that cannot be recovered from that utterance. An ETC system should then satisfy:
\begin{equation}
a\equiv_l b \Rightarrow C(a,l)=C(b,l),
\end{equation}
while preserving the semantics encoded in the speech.
Unlike conventional TN and ITN, which are designed for specific TTS/ASR pre/post-processing, ETC aims to map transcript variants into a consistent canonical representation for fair WER comparison, where either TN or ITN-style transformations can be adopted.
Extending this principle to speech and text translation requires task-specific definitions of admissible variation and semantic-preservation constraints, which we leave to future work.

In this work, our contributions are threefold: (1) we define evaluation-side
transcript canonicalization as a standalone task;
(2) we construct ETC-Bench, comprising 1,098 native-speaker-reviewed
minimal pairs covering 181 phenomenon types across 12 languages,
organized by an extensible taxonomy, and propose Semantic-preserving
Exact-match Normalization Accuracy (SENA); and (3) we release the
Taxonomy-guided Agent Canonicalization Skill (TACS) and assess conventional
and agent-based canonicalization across languages and phenomenon categories.
Agents using TACS achieve higher SENA than the evaluated conventional
baselines on shared languages.

\section{ETC-Bench}
We construct ETC-Bench, a multilingual benchmark for evaluating ETC. 
The benchmark comprises 1,098 minimal pairs covering 181 distinct
canonicalization phenomena across 12 languages: Arabic (ar), Bengali (bn),
German (de), English (en), Persian (fa), French (fr), Hindi (hi), Indonesian
(id), Japanese (ja), Russian (ru), Vietnamese (vi), and Mandarin Chinese (zh).  

\subsection{Phenomenon Taxonomy}
\begin{table}[t]
  \newcommand{\examplepair}[2]{%
    \parbox[c]{\linewidth}{%
      \setlength{\baselineskip}{1.25em}%
      \textit{#1}\\
      \textit{#2}}%
  }

  \centering
  \caption{Representative ETC-Bench pairs. Blue highlights distinctions not
  recoverable from speech.}
  \label{tab:benchmark-examples}

  \small
  \setlength{\tabcolsep}{3pt}
  \renewcommand{\arraystretch}{1.25}

  \begin{tabular}{
    m{1.75cm}
    m{2.45cm}
    m{3.65cm}
  }
    \toprule
    Category & Phenomenon & Example Pair \\
    \midrule

    Numeric
    & Cardinal
    & \examplepair
        {The recipe calls for \textcolor{blue}{2} eggs.}
        {The recipe calls for \textcolor{blue}{two} eggs.}
    \\[1.0em]

    Homophonic
    & \makecell[l]{Homophonic\\ Names}
    & \examplepair
        {\textcolor{blue}{Sara} called me yesterday.}
        {\textcolor{blue}{Sarah} called me yesterday.}
    \\[1.0em]

    Abbreviation
    & \makecell[l]{Context-dependent\\Abbreviation}
    & \examplepair
        {\cyrtext{\textcolor{blue}{Ст.} 5 закона ...}}
        {\cyrtext{\textcolor{blue}{Статья} 5 закона ...}}
    \\[1.0em]

    Orthographic
    & \makecell[l]{Umlaut\\Variants}
    & \examplepair
        {Wir kaufen \textcolor{blue}{Öl}.}
        {Wir kaufen \textcolor{blue}{Oel}.}
    \\[1.0em]

    Homophonic
    & \makecell[l]{Third-person\\Pronoun}
    & \begin{CJK*}{UTF8}{gbsn}
        \examplepair
          {\textcolor{blue}{他}负责那部分任务。}
          {\textcolor{blue}{她}负责那部分任务。}
      \end{CJK*}
    \\

    \bottomrule
  \end{tabular}
\end{table}

We develop a taxonomy of transcript ambiguities by
analyzing the outputs of a multilingual ASR model \cite{li2026mosa}.
The taxonomy organizes variation into five categories
(Fig.~\ref{fig:taxonomy}). A \emph{category} is a coarse-grained class defined by the source of transcript variation, which groups related variation, while a
\emph{phenomenon} specifies a pattern of transcript ambiguity through a natural-language
description and serves as the unit for benchmark construction and analysis.

We group the five categories according to whether their phenomena are language-agnostic or language-specific.
Numeric Expressions and Symbols and Markers contain \emph{language-agnostic}
phenomena: their definitions are shared across languages, although their
applicability and realizations may differ. This inventory adapts and extends
PolyNorm \cite{wong2025polynorm}. In contrast, Abbreviations, Orthographic
Variants, and Homophonic Substitutions contain \emph{language-specific}
phenomena whose definitions depend on a language's lexicon, orthography, or
writing system. 
Homophonic Substitutions specifically covers identically pronounced
expressions whose written distinctions cannot be resolved from speech;
these forms need not have identical written meanings. The hierarchy supports
extension to additional languages and phenomena.

\subsection{Benchmark Construction}
Each benchmark item $i$ consists of a minimal pair of sentences $a_i$ and
$b_i$, annotated with a language $l_i$, a category $c_i$, and a phenomenon
$p_i$. 
The full sentences $a_i$ and $b_i$ share a context and differ only in a target span
instantiating the phenomenon. Language-agnostic pairs use shared templates
localized with language-appropriate contexts and values; language-specific
pairs instantiate phenomenon descriptions with suitable lexical items and
contexts.

We use GPT-5.6-Sol to generate 20 candidate pairs per phenomenon for each applicable language and to filter malformed pairs, followed by manual cleaning.
The candidate pool retains an average of 16.2 pairs per language-indexed phenomenon. We then sample 100 pairs per language, allocating at least one pair to every retained phenomenon to prioritize representative coverage.
Twelve native speakers from Beijing Language and Culture University review
both the validity of the phenomenon definitions and the correctness of the
sampled pairs. Experts refine the inventories and clean the samples, yielding
the final release of 1,098 pairs. The released counts reflect this final
curation rather than the initial 100-pair sampling quota. 

\subsection{Evaluation Metric}
\label{sec:metric}
We propose SENA as the evaluation metric of our benchmark.  It measures the proportion of pairs satisfying both an exact-match condition and a semantic-preservation condition.

Let $C(x,l)$ denote a canonicalization system that produces a canonical text using only
input $x$ and its language $l$. The two transcripts within a benchmark pair are processed independently. Let $s_m$ denote BERTScore $F_1$ under semantic encoder $m$\cite{zhang2020bertscore,tobin2022bertscoreasr}.
Each benchmark pair is reviewed to isolate a distinction for which a STT model lacks distinguishing evidence in speech, not necessarily to assert
identical written semantics (e.g., homophonic gender-marked pronouns).
Its source-side similarity therefore supplies a sample-adaptive reference
that allows the targeted written distinction to be removed:
\begin{align}
t_i &= s_m(a_i,b_i) \\
f_i &= \min_{x\in\{a_i,b_i\}}
       s_m\!\left(x,C(x,l_i)\right)
\label{eq:pair_fidelity}
\end{align}
Here, $t_i$ is the similarity between the two original texts, and $f_i$
is the lower of the two similarities between an input and its canonicalized
output. Both conditions must hold:
\begin{equation}
\operatorname{SENA}=\frac{1}{n}\sum_{i=1}^{n}
\mathbb{I}[C(a_i,l_i)=C(b_i,l_i)]
\mathbb{I}[f_i+\delta\geq t_i],
\label{eq:sena}
\end{equation}
where $n$ is the number of evaluated pairs, $\mathbb{I}$ is an indicator
function, and $\delta=0.1$ is a tolerance on the BERTScore difference
$t_i-f_i$, fixed before the experiments. All SENA results are reported as percentages.
The \emph{exact-match condition} requires identical canonicalized texts.
The \emph{semantic-preservation condition} requires $f_i+\delta\geq t_i$,
using the source pair as a reference for limiting additional semantic loss.

\section{TACS: Taxonomy-Guided Agent Canonicalization Skill}
Rule-based normalizers work well for enumerable patterns such as numbers
and dates, but ETC-Bench also covers context-dependent abbreviations,
orthographic conventions, and homophonic substitutions. Covering these
phenomena with handcrafted rules across languages would require extensive
language-specific implementations. LLMs can draw on contextual and
multilingual knowledge, but this ability alone does not ensure that
independently processed variants receive the same output.

We therefore introduce \textsc{TACS}, a taxonomy-guided skill for LLM
agents. An agent can load relevant guidance as needed to canonicalize each
input independently. The taxonomy describes which distinctions to
recognize. Canonicalization contracts define consistent output forms by
language, category, and phenomenon.
Because the contracts specify targets by phenomenon, TACS can express both TN-style written-to-spoken rewrites and ITN-style spoken-to-written rewrites, as well as canonicalizations beyond their conventional semiotic classes.
The skill also instructs the agent to
make local changes while preserving unrelated text.
Table~\ref{tab:canonical-forms} illustrates canonical forms for the five phenomena in Table~\ref{tab:benchmark-examples}.

\begin{table}[h]
  \newcommand{\tablecell}[1]{%
    \parbox[c]{\linewidth}{%
      \raggedright
      #1%
    }%
  }

  \newcommand{\tablerow}{\\[1.2em]}

  \centering
  \caption{TACS canonical forms for the phenomena illustrated in
  Table~\ref{tab:benchmark-examples}.}
  \label{tab:canonical-forms}

  \footnotesize
  \setlength{\tabcolsep}{3pt}
  \renewcommand{\arraystretch}{1.25}

  \begin{tabular}{
    m{0.4cm}
    m{1.55cm}
    m{2.2cm}
    m{3.8cm}
  }
    \toprule
    LID
    & Category
    & Phenomenon
    & Canonical Form
    \\
    \midrule

    \tablecell{en}
    & \tablecell{Numeric\\Expressions}
    & \tablecell{Cardinal}
    & \tablecell{Quantity in ASCII digits.}
    \tablerow

    \tablecell{en}
    & \tablecell{Homophonic\\Substitutions}
    & \tablecell{Homophonic\\Names}
    & \tablecell{Context-appropriate name in\\
                 broad American English IPA.}
    \tablerow

    \tablecell{ru}
    & \tablecell{Abbreviations}
    & \tablecell{Context-dependent\\Abbreviation}
    & \tablecell{Context-appropriate full\\ Russian form.}
    \tablerow

    \tablecell{de}
    & \tablecell{Orthographic\\Variants}
    & \tablecell{Umlaut Variants}
    & \tablecell{Use \emph{ä, ö, ü} instead of \emph{ae, oe, ue}.}
    \tablerow

    \tablecell{zh}
    & \tablecell{Homophonic\\Substitutions}
    & \tablecell{Third-person\\Pronoun}
    & \tablecell{Broad Mandarin IPA with tone.}
    \\[0.5em]

    \bottomrule
  \end{tabular}
\end{table}

The skill guides the agent to use these resources in two steps.
First, it identifies phenomenon instances using the descriptions and the current utterance, resolving overlapping or competing interpretations. 
Second, it generates local replacements according to the
corresponding contracts.
The instructions require non-overlapping replacements, exact preservation
of text outside the selected spans, and retention of spans with unresolved
ambiguity or no applicable contract. These steps organize the agent's reasoning without prescribing separate model calls or intermediate outputs.

Loading TACS starts with \texttt{SKILL.md}, which specifies batch I/O and global constraints. 
The agent is instructed to solve each record independently and write results to the designated output file, without comparing records or using them as evidence for one another.
The skill contains no benchmark pairs or per-item targets.

\begin{table*}[t]
  \centering
  \caption{Performance comparison across languages in terms of SENA (\%). -- indicates unsupported
  languages. Avg. denotes the macro-average
  across supported languages.}
  \label{tab:results-language}
  \footnotesize
  \begin{tabular}{lccccccccccccc}
    \toprule
    System & ar & bn & de & en & fa & fr & hi
    & id & ja & ru & vi & zh & Avg. \\
    \midrule
    NeMo-ITN & 1.2 & -- & 2.2 & 5.7 & -- & 5.6 & 0.0 & -- & 5.3 & 5.0 & 1.0 & 14.1 & 4.4 \\
    NeMo-TN & 0.0 & -- & 2.1 & 18.2 & -- & 3.4 & 2.0 & -- & 2.6 & 4.0 & 6.1 & 5.4 & 4.9 \\
    FunASR-ITN & -- & -- & 2.2 & 5.7 & -- & 5.6 & -- & 3.1 & 5.3 & 5.0 & 1.0 & 2.2 & 3.7 \\
    FunASR-TN & -- & -- & 2.1 & 19.3 & -- & -- & -- & -- & -- & 4.0 & -- & 12.0 & 9.4 \\
    Whisper Normalizer & 0.0 & 0.0 & 0.0 & 40.9 & 5.0 & 5.6 & 0.0 & 8.2 & 3.9 & 4.0 & 7.1 & 0.0 & 6.2 \\
    \midrule
    No-Skill-GPT
    & 0.0 & 24.0 & 0.0 & 0.0 & 0.0 & 0.0 & 18.0
    & 0.0 & 0.0 & 0.0 & 0.0 & 0.0 & 3.5 \\
    No-Skill-DeepSeek
    & 29.1 & 2.0 & 8.6 & 0.0 & 8.8 & 1.1 & 0.0
    & 0.0 & 0.0 & 6.0 & 0.0 & 0.0 & 4.6 \\
    No-Skill-Qwen
    & 15.1 & 20.8 & 12.9 & 17.4 & 9.2 & 10.1 & 13.3
    & 24.1 & 10.5 & 14.7 & 6.5 & 11.6 & 13.9 \\
    \midrule
    TACS-GPT & 51.9 & 26.5 & 42.7 & 44.7 & 62.1 & 39.7 & 40.0 & 44.2 & 38.6 & 40.3 & 46.6 & 41.7 & 43.3 \\
    TACS-DeepSeek
    & 42.6 & 23.1 & 49.8 & 53.0 & 47.9 & 48.3 & 31.0
    & 41.8 & 42.5 & 48.3 & 55.8 & 40.9 & 43.8 \\
    TACS-Qwen
    & 69.0 & 67.7 & 61.3 & 65.5 & 72.9 & 73.0 & 61.3
    & 75.2 & 64.5 & 75.3 & 78.2 & 62.0 & 68.8 \\
    \bottomrule
  \end{tabular}
\end{table*}

\vspace{-0.5cm}
\section{Experiments}
\subsection{Experimental Setup}
We evaluate five baseline configurations on ETC-Bench: NeMo-ITN and NeMo-TN \cite{zhang2021nemo}, FunASR-ITN and FunASR-TN \cite{gao2023funasr}, and Whisper's normalization module \cite{radford2022whisper}, alongside three TACS-based systems.
We include conventional TN and ITN baseline because ETC is direction-agnostic and both TN and ITN can be regard as an implementation of ETC.
Some commercial STT services, such as Azure Speech, also include post-processing that can resolve some transcript ambiguities. We did not evaluate these services because their post-processing is not available as a standalone text-to-text normalizer.
Performance is measured using SENA as defined in Section~\ref{sec:metric}, with BERTScore computed using the \texttt{xlm-roberta-large} encoder \cite{conneau2020xlmr}.
Whisper uses \texttt{EnglishTextNormalizer} for English and 
\texttt{BasicText}\texttt{Normalizer} for the other 11 languages.
We instantiate the TACS-based agent systems with GPT-5.6-Terra, DeepSeek-V4-Flash
\cite{deepseekai2026deepseekv4}, and Qwen3.8-Flash \cite{qwen3.8}, and
refer to them as TACS-GPT, TACS-DeepSeek, and TACS-Qwen, respectively.
GPT-5.6-Terra is accessed through a third-party API provider, whereas
DeepSeek and Qwen use their official endpoints. 
For each TACS-based system, we conduct three runs on the benchmark and report their mean. 

We use DeepSeek-Harness \cite{deepseek-harness2026}, with shell execution,
file editing and search, skill loading, subagent delegation, and web access enabled. 
An initial prompt defines the task and instructs the agent to load the skill, after which execution proceeds automatically without human intervention.
The two sides of each benchmark item are canonicalized in separate fresh agent sessions with independent workspaces to avoid pairwise context sharing. 

\subsection{Results}
\label{sec:results}
Table~\ref{tab:results-language} reports language-wise performance.
The rule-based systems score 3.7--9.4 on average over their supported
languages. FunASR-TN's 9.4 is averaged over four languages. TACS-GPT, TACS-DeepSeek, and TACS-Qwen achieve macro-average scores of
43.3, 43.8, and 68.8, respectively, across all 12 languages.
Their language-wise scores exceed those of the evaluated conventional
systems wherever both have results.

\begin{table}[t]
  \centering
  \caption{Performance comparison across benchmark categories in terms of SENA (\%).}
  \label{tab:results-category}
  \footnotesize
  \begin{tabular}{lccccc}
    \toprule
    System & Num. & Sym. & Abbr. & Orth. & Homo. \\
    \midrule
    NeMo-ITN & 16.0 & 0.6 & 6.8 & 2.8 & 0.0 \\
    NeMo-TN & 23.7 & 3.2 & 6.8 & 0.0 & 0.0 \\
    FunASR-ITN & 13.3 & 1.4 & 5.1 & 1.7 & 0.0 \\
    FunASR-TN & 32.9 & 5.7 & 8.2 & 0.0 & 0.0 \\
    Whisper Normalizer & 6.7 & 1.0 & 3.8 & 8.1 & 7.0 \\
    \midrule
    No-Skill-GPT & 0.3 & 0.5 & 0.0 & 7.7 & 1.9 \\
    No-Skill-DeepSeek & 0.0 & 0.0 & 0.0 & 8.3 & 5.3 \\
    No-Skill-Qwen & 3.0 & 7.3 & 25.6 & 19.1 & 13.0 \\
    \midrule
    TACS-GPT & 44.1 & 29.1 & 53.4 & 55.7 & 20.0 \\
    TACS-DeepSeek & 50.4 & 27.1 & 56.0 & 58.5 & 12.1 \\
    TACS-Qwen      & 74.4 & 44.3 & 85.9 & 82.3 & 49.0 \\
    \bottomrule
  \end{tabular}
\end{table}

Category-wise results in Table~\ref{tab:results-category} reveal a narrow
success profile for the four TN/ITN systems. Their successful cases are
concentrated in numeric expressions and related forms. Successful orthographic cases also include
number-related variants, such as French hyphenated numbers and Chinese
financial numerals. The four systems score only 0.0--2.8 on orthographic
variants and zero on homophonic substitutions. This pattern is consistent
with their focus on written--spoken conversion within implemented grammar
classes, rather than the broader transcript variation in ETC-Bench.
Whisper shows a different profile. Its English-specific normalizer reaches
40.9 on English, with successful cases including spelling and hyphenation
variants. The basic normalizer obtains modest scores in several other
languages through surface transformations: hyphen removal in French and
Vietnamese, width normalization in Japanese, lowercasing in Indonesian
and Russian, and optional diacritic removal in Persian.
However, producing identical outputs does not always yield SENA credit.
In Bengali and Hindi, destructive removal of informative script marks
also makes some pairs identical, but none of these pairs passes the
semantic-preservation condition.

Across the TACS-based systems, abbreviations and orthographic variants score highest,
while symbols and homophonic substitutions remain more difficult.
TACS-Qwen leads all five categories, but its scores on symbols and
homophonic substitutions still leave substantial room for improvement.
The lower scores on Symbols and Markers and Homophonic Substitutions
mainly reflect failures of exact agreement rather than rejection by
the semantic-preservation condition.   Across three TACS-Qwen runs, the semantic-preservation condition rejected none of the 262 symbol/marker pair evaluations with identical outputs, and only 8 of the 283 homophonic evaluations with identical outputs.

To assess the contribution of TACS, we evaluate the same GPT, DeepSeek,
and Qwen agent configurations without loading it. They score 3.5, 4.6, and
13.9, respectively, compared with 43.3 for TACS-GPT, 43.8 for TACS-DeepSeek,
and 68.8 for TACS-Qwen. 
three agents.
The observed rule-based failures illustrate the roles of the two metric
conditions. Target variants that remain different fail the exact-match
condition. Destructive cleanup can instead produce identical outputs while
failing the semantic-preservation condition, as observed in Bengali and
Hindi. In contrast, transformations such as
\emph{cent-trois}/\emph{cent trois} and \emph{Agung}/\emph{agung}
unify the target variants without substantial additional changes and
receive credit. Thus, a simple transformation can be effective, but identical
outputs alone do not establish successful canonicalization. 
At the pre-fixed $\delta=0.10$, the semantic-preservation condition
rejects all 81 observed Whisper merges that remove informative script
marks in Bengali and Hindi. At $\delta=0.15$, 30.9\% of these merges
receive credit. In one agent run, 99.8\% of merged pairs pass the
condition throughout $\delta\in[0.08,0.12]$.
Table~\ref{tab:etc-asr-examples} illustrates how TACS resolves
transcript variants left distinct by Whisper normalization
on real ASR transcripts\cite{li2026mosa} in evaluation.
\begin{table}[t]
  \centering
  \caption{TACS-DeepSeek unifies transcript variants left distinct by Whisper normalization in two real examples in ASR evaluation.}
  \label{tab:etc-asr-examples}
  \setlength{\tabcolsep}{5pt}
  \renewcommand{\arraystretch}{0.95}
  \setlength{\aboverulesep}{2pt}
  \setlength{\belowrulesep}{2pt}

  \begin{tabular}{@{}l l l@{}}
    \toprule
    LID & Whisper (Hyp. / Ref.) & TACS-DeepSeek \\
    \midrule

    \multirow{2}{*}{de}
      & \textit{und sehr \textcolor{red}{gewi{\ss} da{\ss}}}
      & \multirow{2}{*}{%
          \textit{und sehr \textcolor{blue}{gewiss dass}}} \\
      & \textit{und sehr \textcolor{red}{gewiss dass}}
      & \\[1pt]

    \multirow{2}{*}{en}
      & \textit{thinking of you \textcolor{red}{mabel} and}
      & \multirow{2}{*}{%
          \textit{thinking of you
          \textcolor{blue}{\textipa{/"meIb@l/}} and}} \\
      & \textit{thinking of you \textcolor{red}{mable} and}
      & \\

    \bottomrule
  \end{tabular}
\end{table}

\section{Conclusion}
We define evaluation-side transcript canonicalization as an independent task based on an existing evaluation practice. We introduce ETC-Bench, comprising 1,098 minimal pairs covering 181 phenomenon types across 12 languages, together with an extensible taxonomy and construction framework. Experiments reveal limited coverage in conventional normalizers. TACS shows how the taxonomy and canonicalization contracts can guide LLM agents toward broader coverage, although homophonic substitutions remain challenging. 
We release the benchmark and skill and invite community contributions to extend language coverage, phenomenon inventories, and sample diversity.

\section{compliance with ethical standards}
This is a computational study which releases a new dataset.
Native-speaker reviewers provided linguistic quality control but were
not subjects of the research. No human or animal subjects were studied,
and ethical approval was not required.


\bibliographystyle{IEEEbib_et-al}
\bibliography{strings,refs}

\end{document}